\pdfoutput=1

\documentclass[11pt]{article}
\usepackage{authblk}

\usepackage[symbol]{footmisc}

\usepackage[]{EMNLP2023}

\usepackage{times}
\usepackage{latexsym}

\usepackage{booktabs}
\usepackage{multirow}
\usepackage{float}

\usepackage{hyphenat}

\usepackage{enumerate}

\usepackage{graphicx}

\usepackage[most]{tcolorbox}

\usepackage[T1]{fontenc}

\usepackage[utf8]{inputenc}

\usepackage{microtype}

\usepackage{inconsolata}

\newcommand*{\myfont}{\fontfamily{lmss}\selectfont}
\DeclareTextFontCommand{\textmyfont}{\myfont}
\newcommand*{\fit}{\fontsize{10pt}{5pt}\selectfont}
\newcommand*{\fitaffil}{\fontsize{7.5pt}{5pt}\selectfont}

\title{Domain Terminology Integration into Machine Translation:\\
Leveraging Large Language Models
}

\author[1,2,§]{\fit{Yasmin Moslem}}
\author[3]{\fit{Gianfranco Romani}}
\author[4]{\fit{Mahdi Molaei}}
\author[1,5]{\fit{Rejwanul Haque}}
\author[1,6]{\fit{John D. Kelleher}}
\author[1,2]{\fit{Andy Way}\vspace{-5pt}}

\affil[1]{\fitaffil{ADAPT Centre}\vspace{-1pt}}
\affil[2]{\fitaffil{School of Computing, Dublin City University, Dublin, Ireland}\vspace{-1pt}}
\affil[3]{\fitaffil{Thomson Reuters, Zug, Switzerland}\vspace{-1pt}}
\affil[4]{\fitaffil{Department of Computer Engineering, University of Tabriz, Tabriz, Iran}\vspace{-1pt}}
\affil[5]{\fitaffil{Department of Computing, South East Technological University, Carlow, Ireland}\vspace{-1pt}}
\affil[6]{\fitaffil{Hamilton Institute, Maynooth University, Maynooth, Ireland}}

\begin{document}
\maketitle
\begin{abstract}
\nohyphens{
This paper discusses the methods that we used for our submissions to the WMT 2023 \mbox{Terminology} Shared Task for German-to-English~(DE-EN), English-to-Czech~(EN-CS), and Chinese-to-English~(ZH-EN) language pairs. The task aims to advance machine \mbox{translation} (MT) by challenging participants to develop systems that accurately translate technical terms, ultimately enhancing communication and understanding in specialised domains. To this end, we conduct experiments that utilise large language models (LLMs) for two purposes: generating synthetic bilingual terminology-based data, and post-editing translations generated by an MT model through incorporating pre-approved terms.~Our system employs a four-step process:~(i) using an LLM to generate bilingual synthetic data based on the provided terminology, (ii) fine-tuning a generic encoder-decoder MT model, with a mix of the terminology-based synthetic data generated in the first step and a randomly sampled portion of the original generic training data, (iii) generating translations with the fine-tuned MT model, and (iv) finally, leveraging an LLM for terminology-constrained automatic post-editing of the translations that do not include the required terms. The results demonstrate the effectiveness of our proposed approach in improving the integration of pre-approved terms into translations. The number of terms incorporated into the translations of the blind dataset increases from an average of 36.67\% with the generic model to an average of 72.88\% by the end of the process. In other words, successful utilisation of terms nearly doubles across the three language pairs.
}
\end{abstract}

\vspace{-1.5\baselineskip}

\def\thefootnote{}\footnotemark
\def\thefootnote{§}\footnotetext{Correspondence: \texttt{\scalebox{0.8}[1.0]{first\_name.last\_name@adaptcentre.ie}}}
\def\thefootnote{\arabic{footnote}}

\section{Introduction}

The primary goal of the WMT 2023 Terminology Shared Task is to evaluate the ability of MT systems to accurately translate technical terminology. The task aims to assess the extent to which MT models can utilise additional information regarding the translation of terminology. The shared task requires the participants to provide three translations, one without terms and the others with two individual sets of terms.

There have been several advancements in the area of MT domain adaptation, where an MT model is expected to follow the style and terminology of a certain domain or client \cite{Chu2017-mixed-fine-tuning,Kobus2017-domain-control}. Moreover, some researchers give special focus to terminology while training and fine-tuning MT systems \citep{Dinu2019-TerminologyConstraintsTraining,Hu2019-LexiconInduction,Haque2020-Terminology,Michon2020-Terminology,Nayak2023-Instance}. However, forcing an MT model to adhere to certain terminology at inference time is among the most challenging aspects of MT. Hence, several researchers have investigated approaches to terminology-constrained decoding at translation time \citep{Hokamp2017-ConstrainedDecoding,Hasler2018-TerminologyConstraintsMT,Post2018-FastConstrainedDecoding,Hu2019-ImprovedConstrainedDecoding,Exel2020-TerminologyConstrainedMT}. The goal is to ensure that the MT system can accommodate unseen terminology while retaining translation accuracy and fluency.

Recently, since the emergence of advanced LLMs such as GPT-3~\cite{Brown2020-GPT-3}, BLOOM~\cite{BLOOM2022}, PaLM~\cite{Chowdhery2022-PaLM}, Falcon \cite{Penedo2023-Falcon}, Llama~2 \cite{Touvron2023-Llama2}, and Jais \cite{Sengupta2023-Jais} to mention just a few, researchers have been exploring the capabilities of these models for a number of tasks including MT~\cite{Bawden2023-BLOOM-MT,Hendy2023-LLM-MT,Jiao2023-LLM-MT,Moslem2023-AdaptiveMT,Vilar2023-PaLM-MT}. Some work investigates whether it is possible to utilise LLMs for terminology-constrained MT using a pre-defined glossary~\cite{Moslem2023-AdaptiveMT} or even a dictionary~\cite{Ghazvininejad2023-DictionaryMT}. They found the approach is generally effective in increasing the number of terms used in the translation, even for low-resource languages.

We highlight our key contributions with the systems that we submitted for the WMT 2023 Terminology Shared Task as follows:

\begin{itemize}
    \item \textbf{LLMs for domain-specific data augmentation:} In our previous work \cite{Moslem2022-MT-LM}, we employed LLMs, namely GPT-J \cite{Wang2021-GPT-J} and mGPT \cite{Shliazhko2022-mGPT}, to generate domain-specific datasets based on the target sentences in a small authentic dataset, then generated the source sentences with back-translation \cite{Sennrich2016-BT,Poncelas2019-BT}, and finally fine-tuned an encoder-decoder MT model on this data. In this work, we take a couple of steps forward by instructing an LLM, namely ChatGPT \cite{Brown2020-GPT-3,Ouyang2022-InstructGPT}, to generate terminology-based bilingual synthetic data. In other words, the LLM will generate both the source and target sides of translation pairs, making sure the pre-approved target terms provided by the organisers are used in the translations.
    
    \item \textbf{LLMs for terminology-constrained MT and MT post-editing:} In our previous work, we utilised an LLM for translation and provided it with a list of terms to support in-context learning, which improved adherence to the required terminology at inference time \cite{Moslem2023-AdaptiveMT}. We also investigated whether we could use an LLM for post-editing MT generated by other systems. In this work, we prompt ChatGPT to insert missing terms into translations generated by an encoder-decoder MT system. In other words, if some of the translations generated by a fine-tuned MT model still do not include the terms provided by the organisers, we feed these translations into an LLM, namely ChatGPT, instructing it to incorporate these terms while using the same translation.
\end{itemize}

\section{Method}

In our submissions to the WMT 2023 Terminology Shared Task, we followed these steps:

\begin{enumerate}[(i)]
    \item Generate bilingual synthetic data based on the pre-approved terms, using an LLM, namely ChatGPT.
    \item Fine-tune a generic model, OPUS \cite{Tiedemann2020-OPUS-MT}, on a mix of the terminology-based synthetic data generated in (i) and a randomly sampled portion of the original generic training data.
    \item Generate translations of the dev, test, and blind datasets provided by the organisers with the fine-tuned model from (ii).
    \item Apply terminology-constrained automatic post-editing using ChatGPT to incorporate missing terms into translations that do not yet include the required terminology.
\end{enumerate}

\subsection{Synthetic Data Generation}
\label{sec:data-generation}

We used ChatGPT ``gpt-3.5-turbo''\footnote{The model ``gpt-3.5-turbo'' is a relatively efficient and cost-effective option, so we wanted to understand the quality we can achieve with it.} to generate bilingual sentence pairs, using the terms provided by the organisers. So, given a target term, the model was asked to generate multiple translation pairs, including both the source (e.g. German) and the target (e.g. English). For parameters of ChatGPT's API, we used \emph{top\_p} 1 and \emph{temperature} values 0 and 0.3 to generate diverse outputs.

\smallskip
\begin{tcolorbox}[title={Example prompt: Terminology-based generation}, colbacktitle=gray, fonttitle=\scriptsize, boxrule=0.2pt, right=7pt, left=3pt]
    \begin{scriptsize}
    Please use the ``Federal Ministry of Science'' to generate just 20 numbered sentences in German-English in one Python dictionary format.
    \end{scriptsize}
\end{tcolorbox}
\bigbreak

To filter the generated data, we first removed duplicate sentences from the whole dataset, based on both the source and target. Then, we applied language detection of both sides of the data using \textit{fastText}\footnote{\url{https://fasttext.cc/docs/en/language-identification.html}} and \textit{pycld2}\footnote{\url{https://github.com/aboSamoor/pycld2}} libraries to ensure that the generated sentences were in our desired languages. We excluded any sentences whose scores were below a certain threshold, namely 0.9 for \textit{fastText} and 90 for \textit{pycld2}.

The filtering step removed less than 1\% of the generated data. However, due to computational resource and time limitations, we could not use all the generated data. Table \ref{tab:generated-data} reports the number of generated, filtered, and used translation pairs.

Initially, we only had the development and test datasets, so we used them for the German-to-English language pair. Later, when the organisers released the blind dataset, we used the development, test and blind datasets for the Chinese-to-English and English-to-Czech language pairs.

\begin{table}[htp]
\captionsetup{font=footnotesize,labelfont=footnotesize}
\centering
\begin{small}
\begin{tabular}{@{}llll@{}}
\toprule
\multicolumn{1}{c}{\textbf{Lang}} & \multicolumn{1}{c}{\textbf{Raw}} & \multicolumn{1}{c}{\textbf{Filtered}} & \multicolumn{1}{c}{\textbf{Used}} \\ \midrule
\textbf{DE-EN} & 124,215 & 104,318 & 68,265 \\
\textbf{EN-CS} & 187,471 & 103,797 & 64,218 \\
\textbf{ZH-EN} & 90,538 & 72,695 & 49,001 \\ \bottomrule
\end{tabular}
\end{small}
\caption{Terminology-based bilingual data generated by \mbox{ChatGPT} for fine-tuning the OPUS model}
\label{tab:generated-data}
\end{table}

To assess the quality of the bilingual data generated by ChatGPT, we computed cross-entropy scores \cite{Moore2010-Scoring} of the synthetic translation pairs based on the strong encoder-decoder MT model, NLLB-200 3.3B \cite{NLLB2022}. For scoring, we used CTranslate2\footnote{\url{https://github.com/OpenNMT/CTranslate2}} \cite{Klein2020-Efficient} \emph{score\_batch()} method with the parameters \emph{batch\_type} ``tokens'' and \emph{max\_batch\_size} 2024. We scored each synthetic translation pair generated by ChatGPT, and then calculated the average score for the whole dataset. Computing dual cross-entropy scores according to two inverse translation models trained on clean data is an effective method to evaluate data quality \cite{Junczys-Dowmunt2018-Scoring}. Hence, we computed the scores of both directions of each language pair according to the multilingual MT model NLLB-200 3.3B because both directions are generated by ChatGPT. To produce a baseline for translation quality, we generated the translations of the same datasets using NLLB-200 3.3B for each language direction with \emph{beam\_size} 4, and then scored these translations with the same model. As the scores are in the form of negative log probabilities, we converted them to their exponential equivalents for readability, which are reported in Table \ref{tab:scoring}. It is normal that the model NLLB-200 generates higher scores for its own translations; however, we wanted to know to what extent such scores are comparable to those of ChatGPT's synthetic translation pairs. According to the scores, the German$\leftrightarrow$English language pair had the most comparable quality, followed by Czech$\leftrightarrow$English, and Chinese $\leftrightarrow$English language pairs. 

Among the approaches that can be employed for assessing the quality of synthetic bilingual data is semantic similarity between the two sides of each translation pair (e.g. with mUSE \cite{Yang2020-mUSE}). However, the scoring approach that we previously described and used achieves a similar goal while comparing the quality of the synthetic bilingual data to the translation quality of a strong MT baseline model, namely NLLB-200 3.3B.

\begin{table}[htp]
    \captionsetup{font=footnotesize,labelfont=footnotesize}
    \centering
    \begin{small}
    \begin{tabular}{cccc}
     \textbf{Lang}    &  \textbf{ChatGPT}    &  \textbf{NLLB}  & \textbf{Diff.} \\
     \toprule
     \textbf{DE-EN}    &  0.59    &  0.68   & 0.09 \\
     \textbf{EN-DE}    &  0.56    &  0.64   & 0.08 \\
     \textbf{Avg.}    &  0.58    &  0.66    & 0.08 \\ \midrule
     \textbf{CS-EN}    &  0.58    &  0.70    & 0.12 \\
     \textbf{EN-CS}    &  0.49    &  0.58    & 0.09 \\
     \textbf{Avg.}    &  0.54    &  0.64  & 0.10 \\ \midrule
     \textbf{ZH-EN}    &  0.39    &  0.56   & 0.17 \\
     \textbf{EN-ZH}    &  0.09    &  0.34   & 0.25 \\
     \textbf{Avg.}    &  0.24    &  0.45    & 0.21 \\ \bottomrule
    \end{tabular}
    \end{small}
    \caption{Scores of translation pairs generated by ChatGPT based on the NLLB-200 3.3B model}
    \label{tab:scoring}
\end{table}

\subsection{Fine-tuning}
\label{sec:fine-tuning}

Using the term-based synthetic bilingual data generated in the previous step, we fine-tuned encoder-decoder Transformer-based MT models \cite{Vaswani2017-attention}. In particular, we fine-tuned OPUS MT models, with Hugging Face Transformers.\footnote{\url{https://github.com/huggingface/transformers}} We applied mixed fine-tuning \cite{Chu2017-mixed-fine-tuning}; in other words, we fine-tuned the baseline model with a mix of the terminology-based synthetic data generated from the previous step (cf. Section \ref{sec:data-generation}) and a randomly sampled portion of the original generic data used to train the OPUS baseline model. The numbers of segments taken from the OPUS generic data are as follows: CS: 372,928, DE: 419,881, ZH: 462,780. We over-sampled the synthetic terminology-based data to make it the same size as the used portion of generic data. The fine-tuning parameters are as follows: \emph{train} = 0.9, \emph{val} = 0.1, \emph{batch\_size} = 32, \emph{learning\_rate} = 2e-5, \emph{accumulate\_gradient} = 4, \emph{weight\_decay} = 0.01, \emph{num\_train\_epochs} = 1, \emph{max\_input\_length} = 256, \emph{max\_target\_length} = 256. Finally, we used the fine-tuned model to generate translations for the development, test, and blind sets.

At first glance, the fine-tuning step might look redundant if the LLM can achieve the same translation quality directly, either via zero-shot translation or few-shot in-context learning \cite{Moslem2023-AdaptiveMT}. However, domain-specific or terminology-based knowledge distillation \cite{Treviso2023-Efficient-NLP} from a massive LLM to a compact task-oriented MT model can help boost efficiency at inference time while enhancing domain adaptation and terminology adherence. Obviously, when authentic in-domain data is available, it can be used for fine-tuning instead of synthetic data for domain adaptation of the MT model. In production workflows, only segments that do not meet specific quality criteria are passed to either human or automatic post-editing. Hence, deployment of a model fine-tuned on in-domain data can reduce the number of translations that need post-editing.

\subsection{Terminology-constrained Automatic Post-Editing}
\label{sec:term-ape}

For the shared task, the organisers provided two term sets for each source sentence in the test and blind datasets, and expected the participants to generate two translations that use one term set each. In this step of terminology-constrained automatic post-editing, we aim to refine the translations generated by an MT system by inserting the required terminology. To this end, we checked the translations generated by the fine-tuned model from the previous step (cf. Section \ref{sec:fine-tuning}). For each term set provided for the sentence, if the translation does not include all the terms, we ran this step of terminology insertion into the translation.

This step involves instructing ChatGPT to post-edit the translation by making sure it includes all the terms without changing the rest of the translation. For the API's parameters, we used \emph{top\_p} 1 and \emph{temperature} values 0 and 0.2, and then chose the generation that fixed more terms.

\begin{tcolorbox}[title={Example prompt: Terminology-constrained post-editing}, colbacktitle=gray, fonttitle=\scriptsize, boxrule=0.2pt, right=7pt, left=3pt]
    \begin{scriptsize}
    \begin{itemize}
    \itemindent=-13pt
    \setlength\itemsep{-0.1pt}
    \item[] In the following <tgt\_lang> translation, use the <tgt\_term> to translate
    \item[] the <src\_lang> term <src\_term>, and the...\footnotemark \hspace{2pt}Leave everything else \item[] the same.\textbackslash n\textbackslash n
    \item[] 
    \item[] <src\_lang>: <src\_segment>\textbackslash n
    \item[] <tgt\_lang>: <tgt\_segment>
    \end{itemize}
    \end{scriptsize}
\end{tcolorbox}

\footnotetext{We can add more terms, if needed.}

\section{Evaluation}

To assess the effectiveness of our process, we conducted two types of evaluation: (i) term-level evaluation in order to measure the level of adherence to the required terminology, and (ii) sentence-level evaluation in order to see whether the process \mbox{affected} the quality of the overall translation.

\subsection{Term-level Evaluation}

In Tables \ref{tab:test-used-terms} and \ref{tab:blind-used-terms}, we report the number of terms used in the translations of the test and blind datasets, respectively, in respect to the two term sets provided by the organisers. The results show the effectiveness of our proposed process, increasing the integration of the required terms in the final translations of the blind dataset from an average of 36.67\% with the baseline generic model to an average of 72.88\% after the LLM-based post-editing, across the three language pairs. Interestingly, prompting an LLM to integrate the required terms into the translations generated by a fine-tuned encoder-decoder MT model was more effective than solely using the fine-tuned model.

\addtolength{\tabcolsep}{-2.3pt} 

\begin{table}[H]
\captionsetup{font=footnotesize,labelfont=footnotesize}
\centering
\begin{scriptsize}
\begin{tabular}{ccccccc}
\toprule
\textbf{Lang} & \textbf{System} &   \textbf{Total [1]} &  \textbf{Used [1]} &  \textbf{Total [2]} &  \textbf{Used [2]}  & \textbf{Avg~\%}    \\
\midrule
\multirow{3}{*}{\textbf{DE-EN}}  &    Baseline &           432 &          291 &           317 &          168  &  60.18\\
                        &    Fine-tuned &           432 &          302 &           317 &          165  &  60.98 \\
                        &    Term APE &           432 &          \textbf{397} &           317 &          \textbf{239}  &  \textbf{83.65} \\
                        \midrule
\multirow{3}{*}{\textbf{EN-CS}}  &    Baseline &           550 &          221 &           313 &          139  &  42.30\\
                        &    Fine-tuned &           550 &          135 &           313 &          108  &  29.53 \\
                        &    Term APE &           550 &          \textbf{466} &           313 &          \textbf{283}  &  \textbf{87.57} \\
                        \midrule
\multirow{3}{*}{\textbf{ZH-EN}}  &    Baseline &          1779 &          498 &          1938 &          491  &  26.66\\
                        &    Fine-tuned &          1779 &          854 &          1938 &          570  &  38.71 \\
                        &    Term APE &          1779 &         \textbf{1137} &          1938 &          \textbf{886}  &  \textbf{54.81} \\
                        \midrule
\multirow{3}{*}{\textbf{Avg.~\%}}& Baseline  & & & & &  43.05 \\
                        & Fine-tuned  & & & & & 43.07 \\
                        & Term APE  & & & & & \textbf{75.34} \\ \bottomrule
\end{tabular}
\end{scriptsize}
\caption{For the test dataset, the number of terms used in the translations from the first term set~[1] and the second term set~[2]. According to the results, terminology-constrained automatic post-editing (``Term APE'') using ChatGPT achieved the best adoption of the required terminology.}
\label{tab:test-used-terms}
\end{table}
\label{tab:test}

\begin{table}[H]
\captionsetup{font=footnotesize,labelfont=footnotesize}
\centering
\begin{scriptsize}
\begin{tabular}{ccccccc}
\toprule
\textbf{Lang} & \textbf{System} &   \textbf{Total [1]} &  \textbf{Used [1]} &  \textbf{Total [2]} &  \textbf{Used [2]}  & \textbf{Avg~\%}    \\
\midrule
\multirow{3}{*}{\textbf{DE-EN}}  &    Baseline  &         11357 &         4120 &         11202 &         4623  &  38.77\\
                        &    fine-tuned &         11357 &         4130 &         11202 &         4621  &  38.81\\
                        &    Term APE &         11357 &         \textbf{6257} &         11202 &         \textbf{5893}  &  \textbf{53.85}\\
      \midrule
\multirow{3}{*}{\textbf{EN-CS}}  &    Baseline  &         10626 &         3964 &         10563 &         5122  &  42.90\\
                        &    Fine-tuned &         10626 &         3397 &         10563 &         4412  &  36.87\\
                        &    Term APE &         10626 &         \textbf{8727} &         10563 &         \textbf{8681}  &  \textbf{82.16}\\
      \midrule
\multirow{3}{*}{\textbf{ZH-EN}}  &    Baseline  &          2892 &         1375 &          2908 &          265  &  28.33\\
                        &    Fine-tuned &          2892 &         1422 &          2908 &          970  &  41.26\\
                        &    Term APE &          2892 &         \textbf{2471} &          2908 &         \textbf{2322}  &  \textbf{82.65}\\
\midrule
\multirow{3}{*}{\textbf{Avg.~\%}}&    Baseline  &          &         &          &         &  36.67\\
                        & Fine-tuned  & & & & &38.98\\
                        & Term APE& & & & &\textbf{72.88}\\ \bottomrule
\end{tabular}
\end{scriptsize}
\caption{For the blind dataset, the number of terms used in the translations from the first term set~[1] and the second term set~[2]. According to the results, terminology-based automatic post-editing (``Term APE'') using ChatGPT achieved the best adoption of the required terminology.}
\label{tab:blind-used-terms}
\end{table}

\subsection{Sentence-level Evaluation}

After the end of the submission phase, the organisers released the references for the participants to conduct automatic evaluation. The main purpose of this sentence-based evaluation process is to determine whether terminology integration affected the overall quality of translation. In general, as demonstrated in Table \ref{tab:blind-used-terms} and Table \ref{tab:mt-automatic-evaluation}, this terminology-constrained automatic post-editing step significantly increased the inclusion of the necessary terms into the final translation while improving translation quality across the three language pairs.

For the automatic evaluation of each MT system, we used the BLEU \cite{Papineni2002-BLEU}, chrF++ \cite{Popovic2017-chrF++}, and COMET \cite{Rei2020-COMET} metrics. Since many of the Chinese-to-English segments in the blind dataset did not have two term sets, we evaluated only those that had two term sets (1629 segments out of 2640 segments). We observe that the evaluation scores of the Chinese-to-English translation task are much lower than those of the two other language pairs. This can be due to the literary nature of the blind dataset extracted from Chinese novels, which might be difficult for both the MT model and automatic evaluation metrics.

\begin{table}[H]
\captionsetup{font=footnotesize,labelfont=footnotesize}
\centering
\begin{scriptsize}
\begin{tabular}{ccccccc}
\toprule
\textbf{Lang}   &   \textbf{Count}     &  \textbf{System}      &  \textbf{BLEU}     & \textbf{ chrF++}   &   \textbf{COMET}  \\
\toprule
\multirow{5}{*}{\textbf{DE-EN}} &  \multirow{5}{*}{2963} &  Baseline    &  19.81    &  48.04    &   21.81  \\
                                &        &  Fine-tuned  &  19.27    &  47.75    &   21.51   \\
\cmidrule{3-6}
                                &  &    Term APE [1]    & 32.36     & 60.84     &   40.25   \\
                                &  &    Term APE [2]    & 27.84     & 56.84     &   33.20   \\
                                &   &   Term APE Avg.  &  \textbf{30.10}    &  \textbf{58.84}    &   \textbf{36.73}   \\
\midrule
\multirow{5}{*}{\textbf{EN-CS}}     &   \multirow{5}{*}{3005} &  Baseline   &  29.13  &  53.11  & 50.90   \\
                                    &   &  Fine-tuned         &  24.54      &  49.14  &  33.78    \\
\cmidrule{3-6}
                                    &  &  Term APE [1]  &  45.65    &  67.36    &   79.84    \\
                                    &   &  Term APE [2] &  37.88    &  61.19   &   63.64    \\
                                    &  &  Term APE Avg. & \textbf{41.77}    & \textbf{64.28}    &   \textbf{71.74}   \\
\midrule
\multirow{5}{*}{\textbf{ZH-EN}}  &   \multirow{5}{*}{1629}   &  Baseline  &  6.95 &  27.95 & -50.90 \\
                                 &  &  Fine-tuned     & 7.76  & 29.26     &   -38.83 \\
\cmidrule{3-6}                   &   &  Term APE [1]  &  9.56 &  32.80    &   -18.96  \\
                                 &   &  Term APE [2]  &  11.93    &  35.30    &   -13.51     \\
                                 &  &  Term APE Avg. & \textbf{10.75}     & \textbf{34.05}     & \textbf{-16.24}     \\
\bottomrule
\end{tabular}
\end{scriptsize}
\caption{Automatic evaluation of the overall translation quality across the three language pairs based on the blind dataset. The ``Baseline'' refers to the OPUS model without fine-tuning, while ``Fine-tuned'' refers to the model after domain adaptation with the bilingual terminology-based synthetic data generated by an LLM. Finally, the three last rows for each language pair refer to using ChatGPT for terminology-constrained automatic post-editing (``Term APE'') of the MT output generated by the fine-tuned model. In other words, ``Term APE [1]'' indicates the results when the first term set was used to prompt ChatGPT to integrate terms of this set into the translation generated by the fine-tuned model, while ``Term APE [2]'' refers to using the second term set. Finally, ``Term APE Avg.'' is the average of ``Term APE [1]'' and ``Term APE [2]'' for each language pair. Terminology-constrained automatic post-editing with ChatGPT achieves the best results across the three language pairs in terms of the overall translation quality. As reported in Table \ref{tab:blind-used-terms}, the number of terms integrated after the automatic post-editing step also increased.}
\label{tab:mt-automatic-evaluation}
\end{table}

Moreover, it is worth noting that we used the English term while generating bilingual synthetic data (cf. Section \ref{sec:data-generation}) for the three language pairs. However, English is the target language for both Chinese-to-English and German-to-English language directions, while it is the source language for the English-to-Czech language direction. This can explain the performance degradation after the fine-tuning step in the English-to-Czech language direction (cf. Tables \ref{tab:blind-used-terms} and \ref{tab:mt-automatic-evaluation}). In other words, it is recommended in the step of bilingual synthetic data generation to either use the target term or both the source and target terms while prompting the LLM to generate translation pairs.

As explained in Section \ref{sec:term-ape}, our final step of terminology-constrained automatic post-editing involves instructing an LLM to insert terms that were missing from the output of the fine-tuned model. This significantly increased term usage across all the Chinese-to-English, English-to-Czech, and German-to-English language pairs (cf. Table \ref{tab:blind-used-terms}). Furthermore, as demonstrated in Table \ref{tab:mt-automatic-evaluation}, this step had no detrimental effects on translation quality. In fact, integrating the necessary terms into the translation using ChatGPT improved translation quality according to our automatic evaluation.

\section{Conclusion and Future Work}

In this work, we showed that applying a multistep process of mixed fine-tuning on terminology-based synthetic bilingual data and then terminology-constrained automatic post-editing with an LLM can increase the adherence to the pre-approved terms in the generated translations. By the end of the process, the use of the required terms has increased in the translations of the blind dataset across the three language pairs from an average of 36.67\% with the baseline generic model to an average of 72.88\% after instructing an LLM to integrate the required terms into the translations.

Due to the task restrictions, we had to fine-tune OPUS models only. We would like to experiment with fine-tuning NLLB models, and probably the new SeamlessM4T \cite{Barrault2023-SeamlessM4T}, Mistral \cite{Jiang2023-Mistral}, and MADLAD-400 models \cite{Kudugunta2023-MADLAD}, on the same data and compare the output quality. In our experiments, we employed ChatGPT ``gpt-3.5-turbo'' for both terminology-based synthetic data generation and terminology-constrained automatic post-editing, as it is a relatively efficient and cost-effective option. In the future, we would like to repeat the same experiments with GPT-4 in order to assess the benefit of using a stronger language model on overall performance. We observe that BLOOM can be used as an alternative LLM for data generation; however, one-shot generation might work better than zero-shot generation. In this case, the prompt can consist of a term, a bilingual sentence pair, and then another term. Interestingly, the model will predict a new translation pair including the second term. While certain open-source models such as Llama~2 and Falcon might be employed for the terminology-constrained automatic post-editing step for certain languages, we suspect that they will need fine-tuning before being reliably usable for most languages.

In future work, we will carry out a deeper analysis of the generated synthetic data together with the outputs of the fine-tuned models in order to understand how the properties of the synthetic data affect the fine-tuning results. It is important also to test the same approach for other languages, especially low-resource language pairs.

Moreover, it would be interesting to exclude the fine-tuning step and assess the overall translation quality after LLM-based post-editing. It is possible that domain adaptation through fine-tuning the baseline MT model either on authentic or synthetic data would still be beneficial. It can lead to domain-specific improvements in the overall translation quality that may not be achievable by the baseline model or the terminology-constrained post-editing step. Again, deploying a model fine-tuned on in-domain data into production can enhance terminology adherence in initial translations. As there is no need to send the translations that already include the pre-approved terms to the LLM for terminology-constrained post-editing, this can reduce the number of translations that require post-editing. Such an efficient workflow can allow us to save resources, and minimise latency at inference time. Similarly, there are potential advantages of employing an LLM for post-editing rather than for direct translation. Instead of solely relying on the translation quality of the LLM, quality estimation can be performed to select the best MT model in general or for the current source text segment. \mbox{Ultimately,} only segments that do not meet quality criteria are then passed to the LLM for post-editing.

\section*{Acknowledgements}

This work is supported by the Science Foundation Ireland (SFI) Centre for Research Training in Digitally-Enhanced Reality (d-real) under Grant No. 18/CRT/6224, the ADAPT Centre for \mbox{Digital} Content Technology under SFI's Grant No. 13/RC/2106\_P2, and Microsoft Research.

\bibliography{paperpile}
\bibliographystyle{acl_natbib}

\end{document}